\documentclass[letterpaper, 10 pt, conference]{ieeeconf}  

\IEEEoverridecommandlockouts                              %
\usepackage{graphicx} 
\usepackage{epsfig} 
\usepackage{soul,color} 
\usepackage{xcolor}
\usepackage{comment}
\usepackage{xspace}
\usepackage{mathtools}
\usepackage{booktabs}
\usepackage{multirow}
\usepackage{enumerate}
\usepackage{balance}
\usepackage{gensymb}
\usepackage{tabulary}
\usepackage{dblfloatfix}
\usepackage{layouts}
\usepackage{algorithm}
\usepackage{algpseudocode} 
\usepackage{cite}
\usepackage{bm}
\usepackage{hyperref}
\usepackage{svg}
\usepackage{cite}
\usepackage{caption}
\usepackage{subcaption}
\usepackage{colortbl}
\usepackage{overpic}
\usepackage{rotating}
\usepackage{afterpage}
\usepackage{float}
\usepackage{multirow}
\usepackage{url}
\usepackage{wrapfig}
\usepackage{listings}
\usepackage{adjustbox}
\usepackage{mdwtab}
\usepackage{array}
\usepackage{setspace}
\usepackage{mathdefs}
\usepackage{sonar}
\usepackage{units}
\usepackage{mdframed}

\newtheorem{theo}{Theorem}
\newenvironment{ftheo}
  {\begin{mdframed}\begin{theo}}
  {\end{theo}\end{mdframed}}
  
\usepackage[many]{tcolorbox}

\def\baselinestretch{0.97}\selectfont

\newcolumntype{?}{!{\vrule width 2pt}}
\definecolor{lightgreen}{rgb}{0.8,1,0.8}

\usepackage{caption}
\graphicspath{{../figs/}}

\makeatletter
\newcount\my@repeat@count
\newcommand{\myrepeat}[2]{%
  \begingroup
  \my@repeat@count=\z@
  \@whilenum\my@repeat@count<#1\do{#2\advance\my@repeat@count\@ne}%
  \endgroup
}

\title{\LARGE \bf

Learning Covariances for Estimation with Constrained Bilevel Optimization

}

\author{Mohamad Qadri, Zachary Manchester, and Michael Kaess 
\thanks{$^{1}$M. Qadri, Z. Manchester, and M. Kaess are with the Robotics Institute, Carnegie Mellon University, Pittsburgh, PA 15213, USA.
        {\tt\small \{mqadri, zmanches, kaess\}@andrew.cmu.edu }}%
\thanks{This work was partially supported by Office of Naval Research award N00014-21-1-2482 and National Science Foundation grant IIS-2008279.}%
}

\begin{document}

\maketitle
\thispagestyle{empty}
\pagestyle{empty}





\begin{abstract}
We consider the problem of learning error covariance matrices for robotic state estimation. The convergence of a state estimator to the correct belief over the robot state is  dependent on the proper tuning of noise models.
During inference, these models are used to weigh different blocks of the Jacobian and error vector resulting from linearization and hence, additionally affect the stability and convergence of the non-linear system. We propose a gradient-based method to estimate well-conditioned covariance matrices by formulating the learning process as a constrained bilevel optimization problem over factor graphs. 
We evaluate our method against baselines across a range of simulated and real-world tasks and demonstrate that our technique converges to model estimates that lead to better solutions as evidenced by the improved tracking accuracy on unseen test trajectories.

\end{abstract}

\section{Introduction}
\noindent Robot state estimation is the problem of inferring the state of a robot (a set of geometric or physical quantities such as position, orientation, contact forces, etc.) given sensor measurements. The problem is typically formulated as Maximum a Posteriori (MAP) inference over factor graphs where each node (robot state $\mathbf{x}_i$) is connected to other states via factors (potentials) $\phi_i$ which are distilled from sensor measurements:
\begin{align}
    \mathbf{x}^\text{MAP} = \argmax_\mathbf{x} \prod_i^N \phi_i(\mathbf{x}_i; \theta_i, z_i)
    \label{mainmap}
\end{align} 
The factors typically assume the form:
\begin{align}
 \phi_i \propto  \exp\left( -\frac{1}{2} ||g_i(\mathbf{x}_i) -z_i)||_{\theta_i}^2 \right)   
\end{align}
which leads, after taking the negative log, to the equivalent non-linear least squares objective: 
\begin{align}
    \hat{\mathbf{x}} = \argmin_\mathbf{x} \sum_{i=1}^N \frac{1}{2} ||g_i(\mathbf{x}_i) - z_i ||_{\theta_i}^2
    \label{maineq}
\end{align} 
 where $\mathbf{x}$ are the state variables, $\mathbf{x}_i$ a subset of $\mathbf{x}$, $\mathbf{z} = \{z_i\}$ the sensor measurements, and $g$ the prediction function which maps states onto the sensor's measurement manifold. \textit{Noise Models} $\{\theta_i \}=\boldsymbol{\theta} $  affect the loss landscape (as seen in objective \ref{maineq}) and, as typical in data assimilation procedures \cite{tabeart2018conditioning}, correspond to error covariance matrices.  These parameters dictate the  weight assigned to each measurement which, given an optimal parameter set $\boldsymbol{\theta}^*$, should ideally correlate with the uncertainty of each sensor. Hence, $\{\theta_i \}$ when inaccurately defined will lead to suboptimal solutions. On the other hand, and specifically because each $\theta_i$ is used to scale different error and Jacobian terms after relinearization, the condition number of each $\theta_i$ is correlated with the overall numerical conditioning and stability of problem \ref{maineq}. 
 \begin{figure}[!t]
    \centering
    \includegraphics[width=0.98\columnwidth]{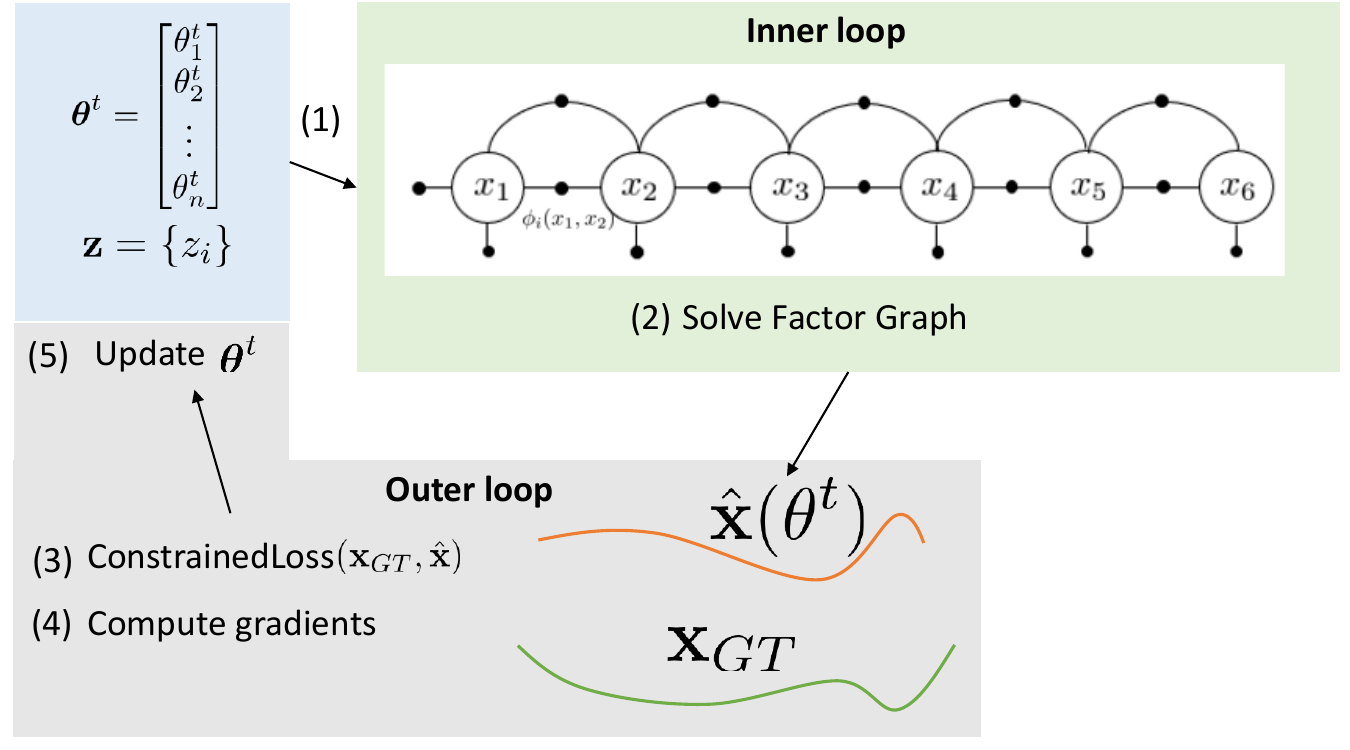}
    \caption{At each iteration, the parameters $\boldsymbol{\theta}^t$ serve as inputs to to the least squares solver. The inner loop optimization outputs a trajectory estimate $\mathbf{\hat{x}}$ which depends on $\boldsymbol{\theta}^t$. The Jacobian of $\mathbf{\hat{x}}$ with respect to $\boldsymbol{\theta}^t$ is computed via numerical differentiation and used to compute the gradient of the loss  with respect to the parameters. Finally, this gradient is  used to update the parameters $\boldsymbol{\theta}$.}
    \label{fig:mainplaceholder}
    \vspace{-6mm}
\end{figure}
\noindent Traditionally the set $\boldsymbol{\theta}$ is manually tuned per application. Nevertheless, alternative approaches exist for estimating $\boldsymbol{\theta}$ from data. Zero-order optimization techniques such as \cite{ hu2015parametric, snoek2012practical, powell1964efficient, nelder1965simplex} can be leveraged but can also quickly become sample inefficient. Other methods attempt to minimize the final tracking error loss by performing gradient-based parameter updates. These techniques generally either 1) rely on unrolling the optimizer which is sensitive to various hyperparameters \cite{yi2021differentiable} or 2) assume that the selected graph optimization algorithm is differentiable \cite{ gradslam}. Note however that this assumption does not hold for state-of-the-art optimizers such as iSAM2 \cite{kaess2012isam2} due to dependence on relinearization thresholds and non-differentiable operations such as removal/insertion of tree nodes. In addition, these methods do not consider the conditioning of the learned parameters. Hence in this work, we make three key contributions:
\begin{itemize}
    \item We formulate the problem as a bilevel optimization problem over factor graphs and use numerical differentiation to efficiently estimate the required gradients.
    \item We propose a technique for estimating well-conditioned  positive definite matrices by incorporating hard condition number constraints into the learning procedure.
    \item We evaluate our approach on different synthetic navigation and real-world planar pushing examples in incremental estimation settings.
\end{itemize}

\vspace{-2mm}
\section{Background and Related Work}
\subsection{Filtering and Smoothing for State Estimation}
\noindent Early state estimation techniques such as the family of Kalman Filters \cite{julier2004unscented, thrun2001robust} rely on the Markov assumption to enable real-time performance. Recent algorithms have been proposed to make these filters differentiable \cite{kloss2021train, sun2016learning, haarnoja2016backprop, lee2020multimodal, jonschkowski2018differentiable}. However, the inherent reliance on the Markov assumption and the inability to re-linearize past states can lead to convergence to poor solutions. Hence, state-of-the-art robotic state estimation algorithms transitioned to factor graph smoothing-based methods which encode the inherent temporal structure avoiding the need to marginalize past states by providing efficient methods to relinearize past estimates \cite{kaess2012isam2, kaess2011bayes, 9982178}. Under Gaussian assumptions, the smoothing problem is equivalent to a non-linear least squares objective weighted by error covariance matrices which often necessitate manual tuning tailored to each application \cite{lambert2019joint, engel2014lsd}.
\subsection{Learned Components in Factor-Graph Inference}
\noindent Recent works incorporate learned components into factor-graph-based inference models. \cite{czarnowski2020deepfactors} learns depth codes which are subsequently used to compute various factors for dense monocular SLAM.  \cite{sodhi2021learning} learns observation models which predict relative sensor poses then used in a factor graph formulation to predict the pose of manipulated objects. Similarly, \cite{baikovitz2021ground} learns a model to predict relative robot poses from non-sequential ground penetrating radar image pairs then used in a factor graph in GPS-denied localization. However, these methods use surrogate losses for learning independently of the graph optimizer or final tracking error  and generally require separate tuning of noise models. 
While traditionally these models have been manually tuned, novel strategies have emerged to learn them directly from data.  \cite{yi2021differentiable, bechtle2021meta} proposed differentiating through the $\argmin$ operator in eq. \ref{maineq} by unrolling the optimizer. However, these techniques are typically sensitive to hyperparameters such as the number of unrolling steps \cite{amos2020differentiable} and, in addition, can suffer from vanishing as well as high bias and variance gradients. Few methods use variational inference techniques to refine noise models when no groundtruth trajectories are available \cite{yoon2021unsupervised, wong2020variational} or use groundtruth trajectories to learn time-correlated measurement noise models \cite{yoon2023towards}. These methods target batch state estimation problems and do not consider the conditioning of the estimated matrices. Recently, a novel method \textit{LEO} \cite{sodhi2022leo} capitalizes on the probabilistic view offered by iSAM2 (as a solver of eq.\ref{mainmap}) to provide a way to learn observation models, by minimizing a novel tracking error. In essence, at every training iteration, LEO samples trajectories from the posterior distribution (a joint Gaussian distribution over the states) and the deviation with respect to the groundtruth trajectory is minimized using an energy-based loss. However, LEO exhibits slow convergence speed due to its dependence on sampling from high-dimensional probability distributions and is prone to convergence to poor local minimas.
\subsection{Covariance Estimation in Mathematical Statistics} 
\noindent The estimation of well-conditioned and stable covariances has garnered considerable attention within the mathematical statistics community as their use spans different statistical methods and practical applications ranging from numerical weather prediction to financial portfolio optimization. \cite{chi2014stable} proposes incorporating a prior involving the nuclear norms of the covariance and its inverse in the estimation process to bound the eigenvalues. \cite{won2013condition} performs maximum likelihood estimation of covariances subject to hard condition number constraints. \cite{tabeart2020improving} proposes new theoretical perspectives on reconditioning covariances using ridge regression or the minimum eigenvalue method. In this work, we propose to estimate error covariance matrices while imposing condition number constraints for the task of incremental robot state estimation.
 
\subsection{Conditioning of Non-linear Least-Squares Problems}
\noindent Iterative methods solve eq. \ref{maineq} by relying on a sequence of linearized subproblems. Each subproblem involves solving the linear system $\mathbf{A}\Delta = \mathbf{b}$ where the stability of the solution is influenced by the condition number of $\mathbf{A}: \kappa(\mathbf{A})$. In fact, 
it has additionally been shown that the convergence rate of specific solvers of the normal equation $\mathbf{A}^T\mathbf{A}\Delta = \mathbf{A}^T\mathbf{b}$, such as conjugate gradient (CG), is upper bounded by $\sqrt{\kappa(\mathbf{A}^T\mathbf{A})}$ \cite{saad2003iterative}. When $\sqrt{\kappa(\mathbf{A}^T\mathbf{A})}$ is high and without proper preconditioning, the performance of CG methods can be especially poor. In section \ref{conditionnumberanalysis}, we show how estimating well-conditioned error covariances matrices is correlated with the stability of the linearized system.

\section{Method}
\noindent Our goal is to learn the parameters $\{\theta_i\} $ using a gradient-based method from groundtruth robots trajectories $\mathbf{x}_{\text{GT}}$. In this work, we view any unconstrained non-linear least squares solver (e.g. Levenberg–Marquardt (LM) or iSAM2), as a function $f(\boldsymbol{\theta}): \mathcal{S}^{n_{1}}_{++} \times \hdots \times \mathcal{S}^{n_{p}}_{++}  \rightarrow \mathcal{X}$ with $\boldsymbol{\theta} = \{\theta_i \; | \; \theta_i \in \mathcal{S}^{n_{i}}_{++}\}$ which, given an initial estimate $\mathbf{x}^0 \in \mathcal{X}: \mathcal{M}_1 \times \hdots \times \mathcal{M}_n$, returns an estimate of the optimal state $\hat{\mathbf{x}} \in \mathcal{X}$ after performing $N$ update steps. Here, $\mathcal{M}_i$ is a Lie Group (e.g. the special Euclidean group $SE(n)$) and $\mathcal{S}^{n_{i}}_{++}$ is the set of $n_i \times n_i$ positive definite matrices. Consider the following inner-outer optimization procedure (also illustrated in Fig. \ref{fig:mainplaceholder}):
%
\begin{align}
    & \text{Inner Loop: } \; f(\boldsymbol{\theta}) =  \argmin_\mathbf{x} H(\mathbf{x},\boldsymbol{\theta}; \mathbf{z}) = \hat{\mathbf{x}}(\boldsymbol{\theta)}  \nonumber \\
    & \quad \quad \quad \quad \quad  \quad \; \; \; \; \, \, = \argmin_\mathbf{x} \sum_i \frac{1}{2}||g_i(\mathbf{x}_i) - z_i||^2_{\theta_i} 
    \label{innerLoop} \\
    & \text{Outer Loop:} \; \min_{\boldsymbol{\theta}} \mathcal{L}(f(\boldsymbol{\theta}), \mathbf{x}_{\text{GT}})
    \label{outerloop}
\end{align}
Where $\mathcal{L}$ is a differentiable loss function capturing the deviation of the estimate $f(\boldsymbol{\theta})$ from the GT. At every iteration, eq. \ref{outerloop} outputs a set $\boldsymbol{\theta}^{t}$. Or, in other words, selects an updated loss landscape for the inner loop optimization such that solving problem \ref{innerLoop} leads to a minima/solution that is closer to the groundtruth trajectory $\mathbf{x}_{\text{GT}}$.
Let $h(\mathbf{x}, \boldsymbol{\theta}) := \frac{\partial H}{\partial \mathbf{x}}$. The graph of $f$ consists of all points satisfying first order-optimality conditions of problem \ref{innerLoop}: $\text{gph}(f) = \{(\boldsymbol{\theta}, f(\boldsymbol{\theta})) \; |  \; f(\boldsymbol{\theta}) = H(\hat{\mathbf{x}}, \boldsymbol{\theta}) \text{ and } h(\hat{\mathbf{x}}, \boldsymbol{\theta} ) =0 \}$. By the chain rule, the gradient $\frac{\partial \mathcal{L}}{\partial \boldsymbol{\theta}}$ requires an estimate of $\frac{\partial f}{\partial \boldsymbol{\theta}}$ (i.e. the Jacobian  of the solution with respect to the parameter vector) and by the implicit function theorem \cite{dontchev2009implicit}, this Jacobian (i.e. $\frac{\partial f}{\partial \boldsymbol{\theta}}$) exists and can be computed   as done in existing work in convex optimization \cite{amos2017optnet, agrawal2019differentiable}. 

\begin{ftheo}
\noindent \textbf{\textit{The Implicit Function Theorem:}}
\newline 
\textit{
Let $\mathbf{\hat{x}}(\boldsymbol{\theta}) := \{\mathbf{x} \; | \; h(\mathbf{x}, \boldsymbol{\theta})=0\}$ where $\mathbf{x} \in \mathcal{X}$ and $\boldsymbol{\theta} = \{ \theta_i  \; | \; \theta_i \in \mathcal{S}^n_{++} \}$. Let $h$ be continuously differentiable in the neighborhood of $(\mathbf{\hat{x}}, \boldsymbol{\theta})$ namely $\frac{\partial h(\mathbf{\hat{x}}(\boldsymbol{\theta}), \boldsymbol{\theta})}{\partial \mathbf{x}}$ be nonsingular. Then:
}
\begin{align}
    \!\! \!\!\frac{\partial f}{\partial \boldsymbol{\theta}} \!=\!\frac{\partial \hat{\mathbf{x}}(\boldsymbol{\theta})}{\partial \boldsymbol{\theta}}\!=\!- \left(\frac{\partial h(\mathbf{\hat{x}}(\boldsymbol{\theta}), \boldsymbol{\theta})}{\partial \mathbf{x}} \right)^{-1} \! \frac{\partial h(\mathbf{\hat{x}}(\boldsymbol{\theta}), \boldsymbol{\theta})}{\partial \mathbf{\boldsymbol{\theta}}}\! 
    \label{gradientIFT}
\end{align}
\end{ftheo}

\noindent Note that the original theorem considers functions operating on vector spaces. However, the theorem can readily be extended to other manifolds by applying the appropriate group operations \cite{pineda2022theseus}. The partial derivatives in eq. \ref{gradientIFT} can be derived and computed analytically. However, since the size of the parameter vector $\boldsymbol{\theta}$ is typically small (first, because each $\theta_i$ is associated with one physical sensor and second, since each $\theta_i$ has a small number of free parameters e.g. a maximum of $6$ for elements residing in $SE(2)$), numerical differentiation proved to be efficient especially when coupled with parallelization on CPU.

\vspace{-2mm}
\subsection{Numerical Jacobians over Lie Groups}

\noindent The left Jacobian of functions acting on manifolds $f: \mathcal{N} \rightarrow \mathcal{M}$ is defined as the linear map from the Lie algebra $T_\mathcal{E}(\mathcal{N})$ of $\mathcal{N}$ to $T_\mathcal{E}(\mathcal{M})$, the Lie algebra of $\mathcal{M}$:
\begin{align}
    & \frac{{}^{\mathcal{E}}Df(\mathcal{Y})}{D\mathcal{Y}} = \lim\limits_{\tau \rightarrow 0 } \frac{f(\tau \oplus \mathcal{Y}) \ominus f(\mathcal{Y})}{\tau}
    \label{mainNumDiff}\\ 
    & \quad \quad \quad \; \; \;  = \lim\limits_{\tau \rightarrow 0 } \frac{\text{Log}(f(\text{Exp}(\tau) \circ \mathcal{Y}) \circ f(\mathcal{Y})^{-1})}{\tau}
\end{align}
where $\mathcal{Y} \in \mathcal{N}$, $\tau$ is a small increment defined on $T_\mathcal{E}(\mathcal{N})$. The Log operator maps elements from a Lie Group to its algebra while the Exp operator maps elements from the algebra to the group.  $\oplus$,  $\ominus$, and $\circ$ are the plus, minus, and composition operators respectively \cite{sola2018micro} where:
\begin{align}
    & \tau \oplus \mathcal{Y} = \text{Exp}(\tau) \circ \mathcal{Y} \\ 
    & \tau = \mathcal{Y}_1 \ominus   \mathcal{Y}_2 = \text{Log}(\mathcal{Y}_1 \circ \mathcal{Y}_2^{-1}); \; \mathcal{Y}_1,\mathcal{Y}_2 \in \mathcal{N}
\end{align}
In this work, $\mathcal{N} = \mathcal{S}^{n_1}_{++} \times \hdots \times \mathcal{S}^{n_p}_{++}$ and $\mathcal{M} = \mathcal{X}$. Additionally, we assume that each vector $\theta_i$ corresponds to the non-zero elements of some corresponding diagonal positive definite matrix $\Sigma_i \in  \mathcal{S}^{n_i}_{++}$. i.e., we define the following map:
\begin{align}
    \theta_i = \text{diag}^{-1}(\Sigma_i) \in \mathbb{R}^{n_i}
\end{align}
where the operator $\text{diag}^{-1}$ constructs a vector from the diagonals of a matrix. Hence, $\tau \in \mathbb{R}^{n_i}$ and the operator $\oplus$ in eq. \ref{mainNumDiff} is the standard addition on vector space $\mathbb{R}^{n_i}$. 
\vspace{-2mm}
\subsection{Constrained Tracking Loss}
\noindent Consider a parameter estimate $\boldsymbol{\theta} \in \mathbb{R}^m$ with $m =\sum_i^p n_i$. Let $\mathcal{D}$ be the training set, $T$ be the total number of states in groundtruth trajectory $\mathbf{x}_{\text{GT}}$, and $D$ be the sum of the Lie algebra dimensions of all states. Then the outer loss is the constrained mean squared error between the estimated trajectory 
and $\mathbf{x}_{\text{GT}}$: 
\begin{align}
& \mathcal{L}(\boldsymbol{\theta})\!=\! 
\frac{1}{2|\mathcal{D}|} \sum_{j=1}^{|\mathcal{D}|} || \textbf{vec}(f(\boldsymbol{\theta})\ominus \mathbf{x}_{\text{GT}})||_2^2  
\label{optprob}\\ 
& \quad \quad \quad \text{subject to: } \boldsymbol{\lambda}^{\text{min}}_i \leq \theta_i \leq \boldsymbol{\lambda}^{\text{max}}_i \quad  \forall \, \theta_i \, \in \, \boldsymbol{\theta} \nonumber
\end{align}
where $\boldsymbol{\lambda}^{\text{min}}_i > 0$ and $\boldsymbol{\lambda}^{\text{max}}_i > \boldsymbol{\lambda}^{\text{min}}_i $ are vectors of minimum and maximum eigenvalues, defined per coordinate and per $\theta_i$, which are enforced to better condition the estimated diagonal matrices as well as ensure their positive definiteness. In other words, these constraints allow to upper bound the condition number $\kappa(\theta_i)=\frac{\boldsymbol{\lambda}^{\text{max}}_i}{\boldsymbol{\lambda}^{\min}_i}$ of the estimated matrices. This, in turn, contributes to better conditioning of the overall linearized system during online inference (see \ref{conditionnumberanalysis}). Hence, since the function $\mathcal{L}(\boldsymbol{\theta})$ is non-convex, the constraints help in steering the optimization towards more desirable minimas. This constrained objective is solved by performing iterative Frank-Wolfe update steps. Algorithm \ref{alg:TrainingLoopMain} outlines a single training iteration. Note that the non-linear least squares (NLLS) in line 3 can be solved by any NLLS optimizer. i.e. our method is agnostic to the choice of optimizer, whether it is differentiable (e.g., Levenberg-Marquardt) or non-differentiable (e.g., iSAM2).
\begin{algorithm}
\caption{Training Loop}\label{alg:TrainingLoopMain}
\begin{algorithmic}[1]
\State \textbf{Input:} Factor Graph $\mathcal{F}$, initialization $\mathbf{x}^0$
\State \textbf{while} itr $<$ max\_iter 
\State \quad $f(\boldsymbol{\theta}^t) = $ Solve[NLLS($\mathcal{F}$, $\mathbf{x}^0$)]
\State \quad Estimate $\frac{\partial \mathcal{L}}{\partial \boldsymbol{\theta}}$ using Eq. \ref{numdiff} 
\State \quad $\boldsymbol{\theta}^{t+1}$ = Frank-Wolfe-Step($\frac{\partial L}{\partial \boldsymbol{\theta}}, \boldsymbol{\theta}^t$)
\State \quad itr = itr+1
\end{algorithmic}
\end{algorithm}
\vspace{-8mm}
\begin{algorithm}
\caption{Frank-Wolfe-Step}\label{alg:FW}
\begin{algorithmic}[1]
\State \textbf{Inputs}: $\frac{\partial \mathcal{L}}{\partial \boldsymbol{\theta}}, \boldsymbol{\theta}^t$
\State Solve (Direction Finding) : 
\State \quad \quad $\mathbf{s}^* = \text{min}_s s^T \frac{\partial \mathcal{L}}{\partial \boldsymbol{\theta}}$ subject to $\boldsymbol{\lambda}^{\text{min}} \leq \mathbf{s} \leq \boldsymbol{\lambda}^{\text{max}} $ 
\State  $\alpha = \frac{2}{M + \text{itr}}$ (Step Size) \Comment{Where M is a parameter}
\State return $ \boldsymbol{\theta}^t + \alpha (\mathbf{s}^* - \boldsymbol{\theta}^t)$ (Updated parameters) 
\label{frankwolfe}
\end{algorithmic}
\end{algorithm}
\newline
The linear program in Alg. \ref{alg:FW} line 3 has negligible computational burden ($m$ decision variables and $2m$ constraints) and can be solved efficiently  using methods such as Interior Point or Dual-Simplex. To estimate, the gradient in Alg. 1 line 4, we propose to perform numerical differentiation by taking advantage of the small parameter space. Taking the gradient of eq. \ref{optprob}, we get:
 \begin{align}
& \frac{\partial \mathcal{L}}{\partial \boldsymbol{\theta}} = \frac{1}{|\mathcal{D}|} \sum_{j=1}^{|\mathcal{D}|}  S(f(\boldsymbol{\theta}))^T
\cdot \underbrace{\textbf{vec}((f(\boldsymbol{\theta})\ominus \mathbf{x}_{\text{GT}})}_{\in \mathbb{R}^{TD}} 
\label{lossmy}
\end{align}

\noindent where $\mathbf{vec}$ is the vectorization operator and $S(f(\boldsymbol{\theta})) \in \mathbb{R}^{TD \times m}$ is a matrix such that each row $r$ is equal to:  
\begin{align}
    &S(f(\boldsymbol{\theta}^t))_{r} = \textbf{vec}\left(\frac{\partial f(\boldsymbol{\theta})}{\partial \theta_{ij}}\right) \\
    & \quad \quad \quad \quad \quad \; \;= \lim\limits_{\tau_{ij} \rightarrow 0} \frac{\textbf{vec}(\text{Log}(f(\boldsymbol{\Tilde{\theta}}) \circ f^{-1}(\boldsymbol{\theta})))}{\tau_{ij}} 
    \label{numdiff}
\end{align}
where $\Tilde{\theta}_{ij} = \theta_{ij} + \tau_{ij}$
 and $\Tilde{\boldsymbol{\theta}} = \boldsymbol{\theta}$ otherwise. By the implicit function theorem, $\frac{\partial f(\boldsymbol{\theta})}{\partial \theta_{ij}}$ exists and is estimated  using finite differencing by perturbing the parameter $\theta_{ij}$ by $\mathbf{\tau}_{ij}$.

\subsection{Remarks} 
\subsubsection{Condition Number Constraints}
\label{conditionnumberanalysis}
\noindent The noise models $\boldsymbol{\theta}$ are used to weight the contribution of the error terms during inference and their eigenvalue spread is correlated with the numerical condition of the linear system resulting from the linearization of eq. \ref{maineq}. Indeed, linearizing eq. \ref{maineq} at iterate $\mathbf{x}^t$, we obtain:
\begin{align}
    \Delta^* = \argmin_{\Delta} \sum_{i=1}^N \frac{1}{2} ||g_i(\mathbf{x}^t_i) + \frac{\partial g_i}{\partial \mathbf{x}_i}\Delta_i - z_i||_{\theta_i}^2
\end{align}
which as shown in \cite{dellaert2017factor} can be re-written as: 
\begin{align}
\!\!\!\!\!\! \! \Delta^*\!\!=\!\argmin_{\Delta} \!\sum_{i=1}^N\!\frac{1}{2}||\theta_i^{-\frac{1}{2}}\frac{\partial g_i}{\partial \mathbf{x}_i}\Delta_i + \theta_i^{-\frac{1}{2}}(g_i(\mathbf{x}_i^t) - z_i)||_{2}^2
\end{align}
Collecting all Jacobians and prediction errors, the system can be rewritten as:
\begin{align}
    \Delta^* = \argmin_{\Delta} \frac{1}{2} ||\mathbf{A}\Delta - \mathbf{b} ||_{2}^2
    \label{mergedLinearSystem}
\end{align}
Since both the Jacobian $\mathbf{A}$ and error vector $\mathbf{b}$ are composed of elements which are matrix multiplied by some $\theta_i^{-\frac{1}{2}}$, the eigenvalue spread $\bar{s}$ which we define as the maximum eigenvalue divided by the minimum eigenvalue over all $\theta_i$:
\begin{align}
\bar{s} = \frac{\max\{  eig(\theta_i) \text{ for } \theta_i \in \boldsymbol{\theta} \} }{\min \{eig(\theta_i) \text{ for }  \theta_i \in \boldsymbol{\theta} \}}
\end{align}
is correlated with the conditioning of matrix $\mathbf{A}$ and the numerical stability of the system in eq. \ref{mergedLinearSystem} \footnote{With factorization-based methods such as QR, poor conditioning  can lead to the loss of significant digits or even yield incorrect solutions.}. 
Our method enables to constrain $\bar{s}$ by incorporating hard constraints into the learning process. 
\subsubsection{Inner Loop Initialization during Training}
\label{Innerloopinit}
We borrow from the imitation learning literature and initialize the inner loop optimizer ($\mathbf{x}^0$ in alg. \ref{alg:TrainingLoopMain} line 3) with the GT trajectory \textit{during training}. These trajectories form our training set and hence, are known a priori. Such initialization has been shown to improve convergence speed and stability \cite{swamy2023inverse} . Note that this is different from baselines such as LEO which, during training, solves the inference problem in eq. \ref{maineq} incrementally  while initializing  $\mathbf{x}^t = \mathbf{x}^{t-1}$ at each new timestep $t$.

\vspace{-1mm}
\section{Results}
\vspace{-3mm}
\subsection{Baselines}
\noindent We compare our method against three baselines: Nelder-Mead \cite{nelder1965simplex}, the Modified Powell's method \cite{powell1964efficient}, and LEO \cite{sodhi2022leo}. \textit{Nelder-Mead} is a gradient-free simplex-based optimization algorithm that aims at decreasing the value of a given function $f$ at the vertices of a working simplex by performing a sequence of transformation (i.e. reflection, shrinkage, etc.).\textit{ Powell's method} is similarly gradient-free and minimizes $f$ by performing a series of one-dimensional line minimization along some search directions. Both methods minimize eq. \ref{outerloop} where $\mathcal{L}$ is the L2 loss. We use SciPy's implementation of these algorithms and run them until convergence.
\textit{LEO} minimizes the following outer-loop energy-based loss: \begin{align}
    &\!\!\!\!\!\!\!\!\mathcal{L}(\theta)\!=\!\frac{1}{|\mathcal{D}|} \! \sum_{(\mathbf{x}_{\text{gt}}^j, \mathbf{z}^j) \in \mathcal{D}}\!\!\!\!\!
E(\boldsymbol{\theta},\mathbf{x}_{\text{gt}}^j;\mathbf{z}^j)\!+\!\log\! \int_\mathbf{x} e^{-E(\boldsymbol{\theta}, \mathbf{x};\mathbf{z}^j)} d\mathbf{x}
\label{leoloss}
\end{align}
where $(\mathbf{x}_{\text{gt}}^j, \mathbf{z}^j)$ is a groundtruth sample  from the training set $\mathcal{D}$, the energy $E( \boldsymbol{\theta}, \mathbf{x}; \mathbf{z}) := H(\boldsymbol{\theta}, \mathbf{x}; \mathbf{z})$ (as defined in eq. \ref{innerLoop}), and the integral is over the space of trajectories. We use the official implementation of LEO by Paloma et al.

\begin{figure*}[b!]
    \includegraphics[width=0.9\textwidth]{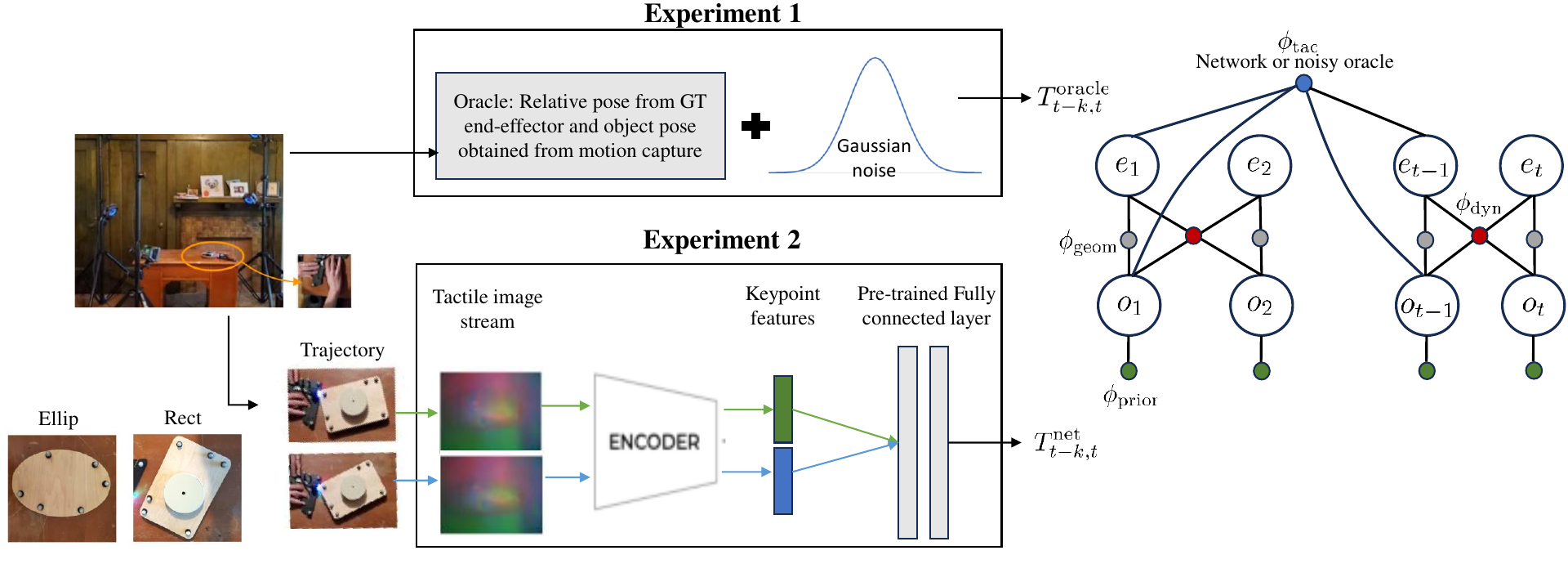}
    \centering
    \caption{Experimental setup and factor graph used for the planar pushing task. Figures showing the data collection setup were borrowed from \cite{sodhi2022leo, sodhi2021learning}}.
    \label{fig:tactilepushfig}
\end{figure*}

\vspace{-1mm}
\subsection{Experimental Details}
\noindent We perform experiments on different simulated and real tasks and use the initialization proposed in \ref{Innerloopinit} for all methods. For all experiments, we start the learning procedure with initial $\boldsymbol{\theta}^0$ values that steer the optimized trajectories far away from the GT or in other words,  $\boldsymbol{\theta}^0$ that are far from the underlying latent unknown parameters.  The implementations of Nelder-Mead and Powell in Scipy allow the specification of bound constraints on the parameters. Hence, we perform 2 sets of experiments where 1) the bounds are loosely specified and effectively only used to ensure the positive definiteness of the estimated matrices  (i.e. $\boldsymbol{\lambda^{\text{min}}}\!>\!\mathbf{0}$) and 2) with tight bounds (denoted with (C) in tables \ref{resnav} and \ref{respush}) where $\lambda^{\text{min}}\!=\! 0.1 $ and $\lambda^{\text{max}}\!=\!10 $ are defined such that the maximum condition number $\kappa^{\text{max}}\!=\!100$ and hence, maximum eigenvalue spread $\bar{s}^{\text{max}
}\!=\!100$. Note that LEO does not support constraints.
After training, the optimal regressed values $\boldsymbol{\theta}^*$ by each method are used as noise models for incremental inference on unseen test samples and the output trajectories are compared against the GT in terms of the root mean square error (RMSE). We use the GTSAM C++ package as the factor graph optimization library and its implementation of iSAM2 as the inner-loop optimizer.
\vspace{-1mm}
\subsection{2D Navigation}
\vspace{-1mm}
 \begin{figure}[H]
    \centering
    \includegraphics[width=0.45\textwidth]{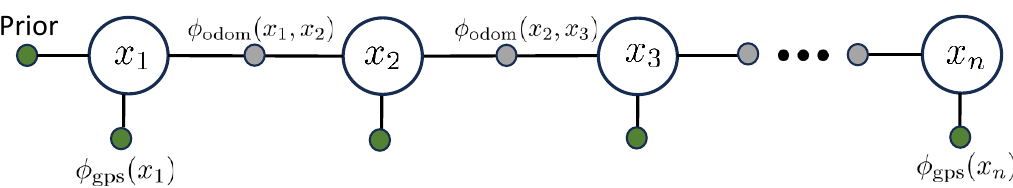}
    \caption{The factor graph used to solve the synthetic robot navigation estimation problems.}
    \label{fig:appendixpic}
    \vspace{-3mm}
\end{figure} 
 \begin{table}[h!]
    \vspace{-2mm}
    \caption{Average RMSE over the testing set for each navigation dataset}
    \begin{center}
    \resizebox{1\linewidth}{!}{
        \begin{tabular}{c|c|c|c|c|c|c}
            \toprule
            \multicolumn{2}{c|}{} & Initial  & LEO   & NMead & Powell & Ours  \\
            \hline
            \multirow{2}{*}{$D_1$} & Transl  &  1.309 &  0.401     & 0.293 & 2.617 & \textbf{0.230}   \\
            & Rot  &  1.184 &  0.070   &   \textbf{0.049} & 0.380 & 0.051  \\
            \hline
             \multirow{2}{*}{$D_2$}&Transl &   2.597 & 1.226 &   1.473  & 7.913  & \textbf{0.775} \\
            &Rot &  2.061 & 0.572  &  0.941 & 0.259 & \textbf{0.056}  \\ 
            \hline
            \multirow{2}{*}{$D_3$}&Transl & 0.918 & 0.640   &  0.482  &  3.960 & \textbf{0.172} \\
            &Rot  &  0.793 &  0.318   &   0.170 &  0.283 & \textbf{0.095} \\ 
            \hline
            \multirow{2}{*}{$D_4$}&Transl & 0.805 & 0.671  &  0.592 &  0.465 &  \textbf{0.209} \\
            &Rot  & 1.123 &  0.816  &    1.092 &  0.212 & \textbf{0.073}\\
            \hline

            \multicolumn{2}{c|}{} & Initial &  & NMead(C)  & Powell (C) &  Ours (C)  \\
            \hline
            \multirow{2}{*}{$D_1$} & Transl  & 1.309 & - & 0.231 & 0.254 & \textbf{0.229}  \\
            & Rot  &  1.184 & -& \textbf{0.048} &  0.050 & 0.051 \\
            \hline
             \multirow{2}{*}{$D_2$}&Transl &   2.597  & - & 1.467 &  0.838 & \textbf{0.777}\\
            &Rot &  2.061 & -& 0.849 & 0.167 &  \textbf{0.051}\\
            \hline
            \multirow{2}{*}{$D_3$}&Transl & 0.918 & -&  0.567 & 0.376 & \textbf{0.173} \\
            &Rot  &  0.793 & -&  0.271 &  \textbf{0.091} & \textbf{0.091}\\ 
            \hline
            \multirow{2}{*}{$D_4$}&Transl & 0.805 &- & 0.681 & 0.356 & \textbf{0.210}\\
            &Rot  & 1.123 & -& 1.090 & \textbf{0.074} &  0.075\\
            \hline
        \end{tabular}
    }
    \end{center}
    \vspace{-6mm}
    \label{resnav}
\end{table}

\noindent We use 4 synthetic planar (i.e. in $SE(2)$) robotic navigation datasets consisting of GPS and odometric measurements all generated from a different set of parameters $\boldsymbol{\theta}^{\text{latent}}$. Datasets $\textit{D}_1$ and $\textit{D}_2$ are generated using fixed parameters $ \{ \theta_{\text{gps}}, \theta_{\text{odom}} 
 \}_{\textit{D}_i} $ while $\textit{D}_3$ and $\textit{D}_4$ use varying parameters which are functions of binary variable $p$: $\{ \theta(p)_{\text{gps}}, \theta(p)_{\text{odom}} 
 \}_{\textit{D}_i} $. $p$ simulates a light detector indicating whether the robot is operating in indoor or outdoor environments.
 We use 5 sample trajectories for training and 20 for testing. Fig.~\ref{fig:appendixpic} shows the structure of the  graph used to estimate the  robot trajectory: A unary GPS factor $\phi_{\text{gps}}(x_i)$ is added to each pose $x_i$ while a binary odometry factor $\phi_{\text{odom}}(x_i, x_{i-1})$ is specified between poses. To simulate realistic robot navigation trajectories, Gaussian noise $\sim \mathcal{N}(0, \theta_{\text{odom}})$ is injected to groundtruth relative odometry measurements while Gaussian noise $\sim \mathcal{N}(0, \theta_{\text{gps}})$ is added to absolute groundtruth GPS measurements. Table \ref{resnav} shows the RMSE of the output trajectories compared against GT. Other than $D_1$ and $D_4$, with constraints enabled, for which Nelder-Mead or Powell generates parameters with slightly better rotation accuracy, our technique consistently converges to parameters $\boldsymbol{\theta}^*$ that lead to better tracking accuracy on all remaining unseen test trajectories. 
\vspace{-1mm}
\subsection{Real-world planar pushing}

\noindent We perform experiments on real-world tactile pushing example where an end-effector pushes an object and the goal is to estimate the pose of both the end-effector and that of the object from sensor data. Groundtruth end-effector and object poses are obtained using a motion capture system. We follow the formulation in \cite{sodhi2021learning} where the graph includes relative pose factor $\phi_{\text{tac}}$ (in which measurements predict the difference in the pose of the end-effector relative to the object between times $t$ and $t-n$ with $n>1$), quasi-static dynamics factor $\phi_{\text{dyn}}$, geometric constraints $\phi_{\text{geo}}$, and end-effector pose priors $\phi_{\text{prior}}$.
Each factor involves a different parameter 
$\in \{\theta_{\text{tac}}, \theta_{\text{dyn}}, \theta_{\text{geo}}, \theta_{\text{prior}}  \} = \boldsymbol{\theta}$ which collectively form our optimization set. We perform two sets of experiments where 1) $\phi_{\text{tac}}$ is provided by a noisy oracle (i.e. simulating an accurate sensor with confident measurements) and 2) $\phi_{\text{tac}}$ being computed from a stream of real tactile measurement. Here, we pre-train a fully connected network on a small training set to output the relative pose measurements from tactile input images. 
\begin{table}[h!]
    \vspace{-2mm}
    \caption{Test RMS translation and rotation errors in cm and rad respectively (averaged over the testing set). We report both end-effector (ee) and object (obj) trajectory error for both the noisy oracle (exp1) and network (exp2) experiments.}
    \begin{center}
    \resizebox{1\linewidth}{!}{
        \begin{tabular}{c|c|c|c|c|c|c|c}
            \toprule
            \multicolumn{8}{c}{\textbf{ELLIP}} \\
            \hline
            \multicolumn{3}{c|}{\textbf{Loose bounds}} & Initial  & LEO & NMead & Powell  &  Ours \\
            \hline
            \multirow{4}{*}{\rotatebox[origin=c]{90}{Noisy Ora}} & \multirow{2}{*}{\rotatebox[origin=c]{90}{ee}} & Transl&   0.784 &0.569 & 0.817 & \textbf{0.012} & \textbf{0.012}  \\
            & & Rot  & 0.004 & 0.041 & 0.004  &  $6e^{-5} $ & $\mathbf{2e^{-5}} $\\
            \cline{2-8}
            & \multirow{2}{*}{\rotatebox[origin=c]{90}{obj}} & Transl& 2.625& 1.875 & 2.581 & 0.510 & \textbf{0.273} \\
            & & Rot  & 0.234 & 0.209 & 0.229 &  0.018 &  \textbf{ 0.008} \\
            \toprule

            \multirow{4}{*}{\rotatebox[origin=c]{90}{Network}} & \multirow{2}{*}{\rotatebox[origin=c]{90}{ee}} & Transl& 0.821 & 0.274 & 0.845 & 0.015 & \textbf{ 0.012} \\
            & & Rot  & $0.004$ & $9e^{-4}$ &  0.004 & $1e^{-4}$ & $\mathbf{1e^{-5}}$ \\
            \cline{2-8}
            & \multirow{2}{*}{\rotatebox[origin=c]{90}{obj}} & Transl&  2.431  & 2.456 & 2.477 & 1.543 &  \textbf{0.982} \\ 
            & & Rot  & 0.271 &  0.162 & 0.265 & 0.168 & \textbf{0.155} \\
            \toprule

                        
           \multicolumn{3}{c|}{\textbf{Tight bounds}} & Initial  &  & NMead (C) & Powell (C)  &  Ours (C) \\
            \hline
            \multirow{4}{*}{\rotatebox[origin=c]{90}{Noisy Ora}} & \multirow{2}{*}{\rotatebox[origin=c]{90}{ee}} & Transl&   0.784 & - & 0.710 & \textbf{0.006} &  0.018  \\
            & & Rot  & 0.004 & -  & 0.005 & $\mathbf{2e^{-5}}$& $\mathbf{2e^{-5}}$\\
            \cline{2-8}
            & \multirow{2}{*}{\rotatebox[origin=c]{90}{obj}} & Transl& 2.625& - & 1.534 & 0.364 & \textbf{0.287} \\
            & & Rot  & 0.234 & - & 0.031 & 0.023 & \textbf{0.007} \\ 
            \toprule

            \multirow{4}{*}{\rotatebox[origin=c]{90}{Network}} & \multirow{2}{*}{\rotatebox[origin=c]{90}{ee}} & Transl& 0.821  &  - & 0.900 &  0.010 & $\mathbf{1e^{-4}}$ \\
            & & Rot  & $0.004$ & - &  0.007  & $\mathbf{2e^{-5}}$ & $7e^{-5}$\\
            \cline{2-8}
            & \multirow{2}{*}{\rotatebox[origin=c]{90}{obj}} & Transl&  2.431  & - & 1.813 & 1.521 & \textbf{0.980} \\ 
            & & Rot  & 0.271 & - & 0.159 &  0.159 & \textbf{0.141}  \\
            \toprule

            \multicolumn{8}{c}{\textbf{RECT}} \\
            \hline
            \multicolumn{3}{c|}{\textbf{Loose bounds}} & Initial  & LEO & NMead & Powell  &  Ours \\
            \hline
            \multirow{4}{*}{\rotatebox[origin=c]{90}{Noisy Ora}} & \multirow{2}{*}{\rotatebox[origin=c]{90}{ee}} & Transl&  1.027 & 1.381 & 0.912 & 0.009 & $\mathbf{8e^{-4}}$   \\
            & & Rot  & 0.006  & 0.015 & 0.006 & $\mathbf{8e^{-5}}$ & 0.007 \\ 
            \cline{2-8}
            & \multirow{2}{*}{\rotatebox[origin=c]{90}{obj}} & Transl&  5.571  & 5.724 & 4.308 & 0.575  &  \textbf{0.396} \\
            & & Rot  &  0.471 & 0.197 &  0.464 & 0.007 &  \textbf{0.009} \\
            \toprule
            \multirow{4}{*}{\rotatebox[origin=c]{90}{Network}} & \multirow{2}{*}{\rotatebox[origin=c]{90}{ee}} & Transl&  0.729  & 2.887 & 0.862 & 0.041 &  \textbf{0.014} \\
            & & Rot  &  0.004 &  0.001 & 0.004 & $2e^{-4}$ & $\mathbf{6e^{-5}}$\\    
            \cline{2-8}
            & \multirow{2}{*}{\rotatebox[origin=c]{90}{obj}} & Transl&  4.300  & 4.935 & 6.464 &  2.505  & \textbf{1.609}  \\
            & & Rot  & 0.529 & 0.287 & 0.556 & 0.229  & \textbf{0.212} \\
            \hline

            \multicolumn{3}{c|}{\textbf{Tight bounds}} & Initial  &  & NMead (C) & Powell (C) &   Ours (C) \\
            \hline
            \multirow{4}{*}{\rotatebox[origin=c]{90}{Noisy Ora}} & \multirow{2}{*}{\rotatebox[origin=c]{90}{ee}} & Transl&  1.027 & - & 1.643 & \textbf{0.007} & 0.019  \\
            & & Rot  & 0.006  & - & 0.016 & $3e^{-5}$   &  $\mathbf{1e^{-5}}$ \\
            \cline{2-8}
            & \multirow{2}{*}{\rotatebox[origin=c]{90}{obj}} & Transl&  5.571  & - & 2.993 & 0.484 & \textbf{0.413} \\
            & & Rot  &  0.471 & - &  0.061 & 0.026  &  \textbf{0.013} \\
            \toprule
            \multirow{4}{*}{\rotatebox[origin=c]{90}{Network}} & \multirow{2}{*}{\rotatebox[origin=c]{90}{ee}} & Transl&  0.729  & - & 1.293 &  0.026 &   \textbf{0.014} \\
            & & Rot  &  0.004 & - &  0.012  & $\mathbf{5e^{-5}}$ & $1e^{-4}$\\
            \cline{2-8}
            & \multirow{2}{*}{\rotatebox[origin=c]{90}{obj}} & Transl&  4.300  & - & 4.229 & 1.645  & \textbf{1.609} \\
            & & Rot  & 0.529 &  - &   0.240 & 0.214 & \textbf{0.207}  \\ 
            \toprule
        \end{tabular}
    }
    \label{respush}
    \end{center}
    \vspace{-3mm}
\end{table}
These measurements are designed to be relatively noisy and inaccurate. Both experiments are performed on 2 objects of different shapes: ellipsoidal and rectangular each featuring different contact patches. Finally, we initialize all  $\theta_i \in \boldsymbol{\theta}^0$ such that 1) they are far from the underlying unknown latents and 2) the spread $\bar{s}$ is  high. We use a 5/10 train/test split.
The experimental setup is further illustrated in Fig.~\ref{fig:tactilepushfig}. Table \ref{respush} shows the RMSE of the output trajectories when compared against GT for the pushing task: LEO can converge to poor local minimas  which lead at times to worse tracking error on the testing set compared to our initial estimate. Similarly, for Nelder-Mead, the method can fail to decrease the function value. In fact, it has practically been observed to stagnate at non-optimal points \cite{nocedal1999numerical}. While Powell's method generates reasonable parameter estimates, our technique exploits the structure of the gradient and converges to solutions $\boldsymbol{\theta}^*$ that lead to better or comparable tracking accuracy on all unseen test trajectories across the different experiments.

\begin{table}[h!]
    \vspace{-2mm}
    \caption{The spread $\bar{s}$ (to the nearest integer) of the final output values $\boldsymbol{\theta}^*$ learned by our method.}
    \begin{center}
    \resizebox{0.9\linewidth}{!}{
        \begin{tabular}{c|c|c|c|c}
            \toprule
            \multicolumn{1}{c|}{}  & Ellip Ora & Ellip Net & Rect Ora & Rect Net\\
            \hline
            Ours (Loose bounds)  & 610 & 3615 & 460 & 420 \\
            Ours (Tight bounds) & 70 & 62 & 67 & 75 \\         
 
            \toprule
        \end{tabular}
    }
    \label{table:spread}
    \end{center}
    \vspace{-3mm}
\end{table}
\noindent Table \ref{table:spread} shows the eigenvalue value spread $\bar{s}$ of the optimized set $\boldsymbol{\theta}^*$ estimated by our method under both tight and loose eigenvalue bounds. Note that the generated solution indeed satisfies the hard bound constraints ($\bar{s}^{\text{max}} < 100$) when specified. Conversely, in the absence of upper-bound constraints, the eigenvalue spread can effectively grow unbounded.
 
\vspace{-1mm}
\section{Commentary }
\subsection{Invariance to the Specified Constraint Bounds}

\noindent We note from Tables \ref{resnav} and \ref{respush}, that the performance of the Nelder-Mead and Powell's algorithms, in terms of tracking accuracy and variance of the output 
is influenced by the specified bound constraints. In contrast, our algorithm converges to minimas that lead to similar tracking performance (albeit with different eigenvalue spread) regardless of the specified bounds. Note that the parameter $M$ in alg \ref{alg:FW} requires tuning in order to damp the step size if the bound interval is large. In addition, and as typical in optimization problems, a solution needs to exist in the feasible region. 
\vspace{-1mm}
\subsection{Runtime}

\begin{figure}[h!]
    \vspace{-3mm}
    \centering
    \includegraphics[width=0.9\columnwidth]{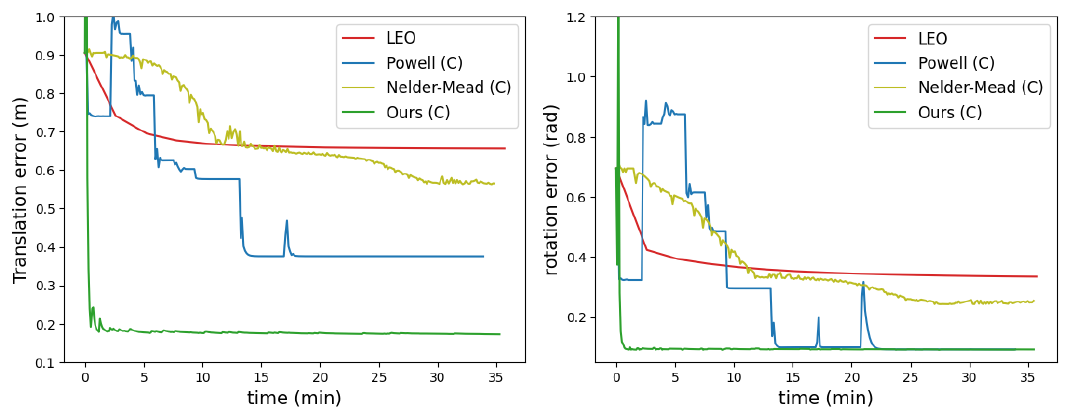}
    \caption{Training translation and rotation error vs runtime for all methods on the navigation $D_3$ dataset.}
    \label{fig:runtime}
\end{figure}

\noindent Fig.~\ref{fig:runtime} shows the translation and rotation accuracy on the training set as a function of runtime for the navigation dataset $D_3$. Nelder-Mead and Powell's method, being zero-order methods, exhibit slower convergence rates compared to gradient-based optimizers.
LEO does leverage the gradient of the energy-based loss (eq. \ref{leoloss}). However, it requires samples from a high dimensional probability distribution to approximate the integral term at each training iteration which is a time-consuming process. Our technique generates gradients by directly comparing deviations from the training trajectories leading to faster convergence. However, note that our method's running  time is expected to increase proportionally to the dimension of $\boldsymbol{\theta}$ since although parallelizable, the perturbations in eq. \ref{numdiff} need to be performed per parameter and per dimension.

\vspace{-4mm}
\subsection{Varying the Initialization}

\begin{figure}[h!]
    \centering
    \includegraphics[width=0.9\columnwidth]{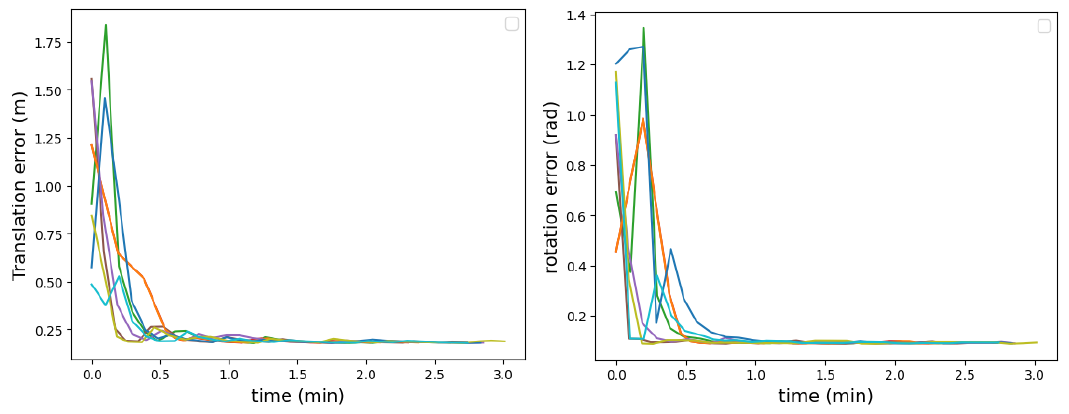}
    \caption{Convergence with varying parameter initialization $\boldsymbol{\theta}^0$}
    \label{fig:changeinit}
\end{figure}

\noindent We show in Fig.~\ref{fig:changeinit} the training error curves for different initializations of the parameter vector $\boldsymbol{\theta}^0$ for dataset $D_3$. We observe that across all initializations, our method converges to model estimates that minimize the translation and rotation error (deviation from groundtruth pose) on the training set.

\vspace{-3mm}

\subsection{Varying the number of training trajectories} 
\noindent We used 5 training samples across experiments and found that increasing the number of training trajectories does not lead to improved generalization and testing accuracy. As noted in \cite{qadri2023learning}, given the relatively small parameter set $\boldsymbol{\theta}$ and the fact that both train and test trajectories are sampled from the same distribution, the learning process only requires a few samples.

\section{Conclusion and Future Work}
\noindent We introduce a gradient-based algorithm to learn error covariance matrices for robotic state estimation. Our technique formulates the problem as a bilevel optimization procedure and generates required gradients through numerical differentiation.  Our method results in parameters that generalize better compared to baselines with the added benefit of incorporating hard condition number constraints. In future work, we want to extend our algorithm to learn parameters $\{\theta_i\}$ that are themselves functions of observations i.e. $\theta_i(z_i, \Theta_i)$ where $\Theta_i$ can, for example, be the weights of a jointly trained neural network. Indeed, the outputs of the network can be perturbed to approximate $\frac{\partial\mathbf{\hat{x}}}{\partial \theta_i}$ (where $\mathbf{\hat{x}}$ is the solution returned by the graph optimizer) as proposed in this work and then chained with $\frac{\partial \theta_i}{\partial \Theta}$ (obtainable from existing auto-differentiation packages such as PyTorch \cite{paszke2017automatic}) to get the Jacobian of the optimized output trajectory with respect to network weights. Finally, enforcing constraints on the output of a neural network offers interesting related avenues for future research.

\bibliographystyle{IEEEtran}
\IEEEtriggeratref{21}
\bibliography{main}

\end{document}